\documentclass{article}

\PassOptionsToPackage{numbers, compress}{natbib}

\usepackage[preprint]{neurips_2025}

\usepackage[utf8]{inputenc} 
\usepackage[T1]{fontenc}    
\usepackage{hyperref}       
\usepackage{url}            
\usepackage{booktabs}       
\usepackage{amsfonts}       
\usepackage{nicefrac}       
\usepackage{microtype}      
\usepackage{xcolor}         
\usepackage{amssymb}
\usepackage{breqn}
\usepackage{mathtools}
\usepackage{subcaption}
\usepackage{hyperref}
\usepackage{algorithm}
\usepackage{algpseudocode}
\usepackage{enumitem}
\usepackage{pifont}
\usepackage[absolute,overlay]{textpos}
\usepackage{soul}
\usepackage{booktabs}  
\usepackage{array}     
\usepackage{animate}
\usepackage{wrapfig}

\input{mysymbol.sty}

\title{Model-Driven Graph Contrastive Learning}

\author{%
  Ali Azizpour\\
  Rice University\\
  \texttt{ali.azizpour@rice.edu}\\
  \And
  Nicolas Zilberstein\\
  Rice University\\
  \texttt{nzilberstein@rice.edu}\\
  \And
  Santiago Segarra\\
  Rice University\\
  \texttt{segarra@rice.edu} \\
}

\begin{document}

\maketitle

\begin{abstract}
We propose \textbf{MGCL}, a model-driven graph contrastive learning (GCL) framework that leverages \emph{graphons} (probabilistic generative models for graphs) to guide contrastive learning by accounting for the data's underlying generative process.
GCL has emerged as a powerful self-supervised framework for learning expressive node or graph representations without relying on annotated labels, which are often scarce in real-world data.
By contrasting augmented views of graph data, GCL has demonstrated strong performance across various downstream tasks, such as node and graph classification.
However, existing methods typically rely on manually designed or heuristic augmentation strategies that are not tailored to the underlying data distribution and operate at the individual graph level, ignoring similarities among graphs generated from the same model.
Conversely, in our proposed approach, MGCL first estimates the graphon associated with the observed data and then defines a graphon-informed augmentation process, enabling \emph{data-adaptive and principled augmentations}. 
Additionally, for graph-level tasks, MGCL clusters the dataset and estimates a graphon per group, enabling contrastive pairs to reflect shared semantics and structure.
Extensive experiments on benchmark datasets demonstrate that MGCL achieves state-of-the-art performance, highlighting the advantages of incorporating generative models into GCL.
\end{abstract}

\section{Introduction}

Graph Neural Networks (GNNs)~\cite{gnn, chien2024opportunities} have achieved remarkable success across a wide range of domains, including wireless networks~\cite{zhao2022link, chowdhury2021unfolding, netvigil}, bioinformatics~\cite{komb,grassrep}, and social networks~\cite{kumar2022influence}. 
Their ability to capture and propagate structural information over graphs has made them powerful tools for learning node and graph-level representations~\cite{xu2018representation,ying2018hierarchical}. 
However, a key limitation of GNNs is their reliance on task-specific supervision. 
To learn discriminative representations, GNNs typically require labeled data, limiting their applicability in real-world settings where annotations are scarce or costly~\cite{liu2022graph,hu2019strategies}. 
To address this issue, graph contrast learning (GCL) has recently emerged as a promising self-supervised alternative that enables \emph{representation learning on graphs without requiring labels}~\cite{dgi, graphcl, grace, gca, mvgrl, sgrl}.

GCL aims to learn informative node or graph representations (based on the task of interest) by maximizing the agreement between different augmented views of the same graph while minimizing the agreement with the views of other graphs or corrupted versions~\cite{graphcl, gca}. 
This enables the model to capture essential structures in the absence of labeled data, and has proven effective for a variety of downstream tasks, including node classification, clustering, and graph classification~\cite{gouda, grade, joao}.

Despite their effectiveness, existing GCL methods often rely on manually designed augmentation strategies, such as node or edge dropping~\cite{rong2020dropedge, graphcl}, feature masking~\cite{grace, gca}, or subgraph sampling~\cite{graphcl}, which are heuristic and task-specific~\cite{joao}, limiting their adaptability across many graph structures and tasks. 
Recent advances have introduced learnable augmentations using prior knowledge or gradient-based feedback (e.g., spectral perturbations or adversarial training~\cite{ad-gcl, gpa, liu2022revisiting, markCL}), but these approaches still explore a limited augmentation space.
Additionally, many GCL frameworks operate at the individual graph level and treat all other samples as negative samples, which can lead to false negative pairs when structurally or semantically similar graphs are contrasted~\cite{graphcl, ad-gcl, sgrl}, ultimately degrading representation quality.

\begin{wrapfigure}{r}{0.40\textwidth}  
    \centering
    \vspace{-2mm}
    \includegraphics[width=0.39\textwidth]{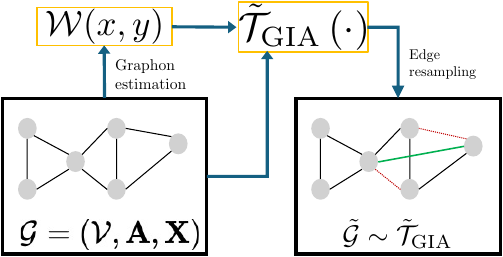}
    \caption{\small{Using the underlying graphon to inform the augmentation.}}
    \vspace{-2mm}
    \label{fig:highlevel}
\end{wrapfigure}

\begin{figure*}[t]
	\centering
	\includegraphics[width=\textwidth]{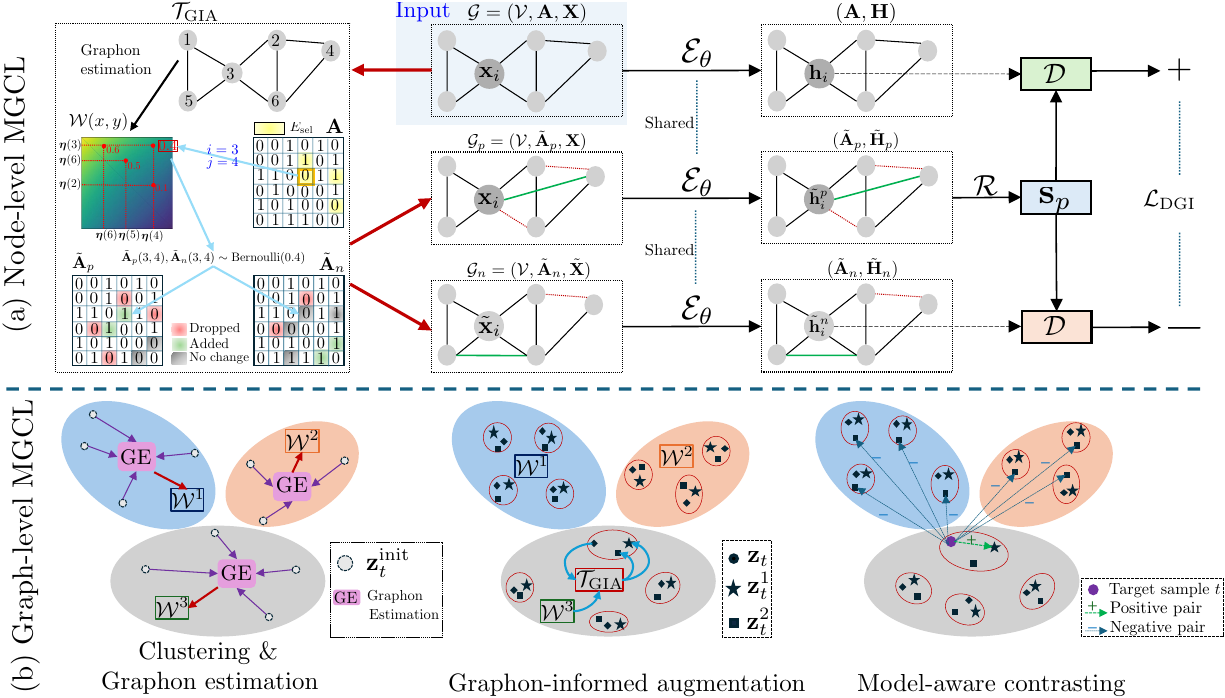}
	\caption{\small{An overview of MGCL. 
        (a)~For node-level tasks, MGCL estimates the underlying graphon to guide augmentation. Two views are sampled from \(\ccalT_{\text{GIA}}\), with the edge between nodes \(3\) and \(4\) highlighted as an example. 
        (b)~For a set of graphs, MGCL first clusters the graphs (each white point) and estimates a graphon for each cluster. Two augmentations are then generated for each graph based on its corresponding graphon. Finally, each graph is pulled toward its own first view while being pushed away from the second views of graphs in other clusters, enabling model-aware contrastive learning.}
        }
	\label{fig:overview}
\end{figure*}

In this work, we propose \textbf{MGCL}, a \textbf{M}odel-driven \textbf{G}raph \textbf{C}ontrastive \textbf{L}earning framework that explicitly incorporates the underlying generative process into contrastive learning.
MGCL assumes that graphs are samples from a shared, but unknown, graphon -- a nonparametric probabilistic model for generating graphs~\cite{graphon, diaconis2007graph, gmixup}, which has shown success in various applications~\cite{parise2023graphon, navarro2022joint, gao2019graphon, roddenberry2021network, navarro2023graphmad}.
We leverage this assumption to construct Graphon-Informed Augmentations (GIAs): data-driven stochastic transformations that generate semantically faithful views guided by the estimated graphon, as illustrated in Figure~\ref{fig:highlevel}.
MGCL supports both node and graph-level tasks; for the latter, it clusters the dataset into structurally similar groups and estimates a separate graphon per cluster.
Through the identification of common structural patterns, MGCL \emph{reduces false negatives}, a key challenge in graph-level contrastive learning. 
A full overview is shown in Figure~\ref{fig:overview}.

We validate the advantages of MGCL through extensive experiments on different real-world datasets, which demonstrate that MGCL achieves state-of-the-art performance on node and graph classification tasks, highlighting the advantages of incorporating generative models into GCL.

In summary, our main contributions are as follows:  
\begin{itemize}
    \item We introduce \textbf{MGCL}, the first framework to incorporate \emph{graphons} for \emph{model-based augmentations} in GCL.
    \item For node-level tasks, we use a single graphon to construct the GIAs, capturing common structural patterns across the dataset.
    \item For graph-level tasks, we introduce a novel approach that partitions the dataset into groups of structurally and semantically similar graphs, assigning a dedicated graphon to each group for GIA construction and model-aware contrastive learning.
    \item We conduct extensive experiments on public benchmark datasets and tasks, demonstrating that MGCL outperforms state-of-the-art baselines.
\end{itemize}

\section{Background and related works}

\subsection{Graph contrastive learning}
\label{subsec:background_gcl}
GCL aims to learn discriminative node or graph-level embeddings in a self-supervised manner, without relying on explicit labels. 
Given a graph $\ccalG=(\ccalV, \ccalE)$ (or a collection of graphs $\{\ccalG_t\}_{t=1}^T$) with $|\ccalV|=N$ nodes, the goal of GCL is to train an encoder \(\ccalE_{\theta}(\cdot)\) so that it produces expressive representations or embeddings. 
Broadly, there are two types of augmentation: node-level and graph-level.

\paragraph{Node-level.} 
In the node-level case, given an attribute matrix \(\bbX \in \reals^{N\times F}\) and an adjacency matrix \(\bbA\in \{0,1\}^{N\times N}\), the encoder learns to generate embeddings \(\bbH = \ccalE_{\theta}({\bbA}, \bbX)\), where \(\bbH \in \mathbb{R}^{N \times K}\), which can then be used for various downstream tasks.
One widely used approach for node-level contrastive learning is based on DGI~\cite{dgi}, which maximizes mutual information between local node embeddings and a global summary vector of the graph.  
Specifically, DGI encourages the embedding of a target node \(i\) from the original view, \(\mathbf{h}_i = [\mathbf{H}]_{i,:}\), to align with the graph-level summary vector \(\mathbf{s} = \mathcal{R}(\mathcal{E}_\theta(\mathbf{A},\mathbf{X}))\), where \(\mathcal{R}\) is a read-out function. 
Simultaneously, it ensures that embeddings from a corrupted view, \(\tilde{\mathbf{h}}_i\), often created by feature shuffling, are pushed away from this summary.
GraphCL~\cite{graphcl} extends the DGI framework by introducing augmentations of the input graph through predefined perturbations (e.g., random edge drop or feature masking) applied to the adjacency matrix \(\bbA\).  
The graph-level summary vector is then obtained from the augmented version of the graph; this strategy has been shown to improve performance.
Still, it relies on hand-crafted augmentations.
Another widely used family of methods in this context are \emph{InfoNCE-based} approaches~\cite{infoNCE}.  
These methods generate two augmented views of the input graph, which are encoded by the encoder \(\ccalE_\theta(\cdot)\) to produce node embeddings for each view.  
For a given target node \(i\), its representation in one view (\(\bbh_i\)) is trained to be similar to that in the other view (\(\tbh_i\), \emph{the positive pair}), while being dissimilar to representations of all other nodes (\(\tbh_j\), for all \(j \neq i\), that is, \emph{the negative pairs})~\cite{gca,grace, mvgrl}.

\paragraph{Graph-level.}
In graph-level contrastive learning, the encoder outputs an embedding per graph $\ccalG_t$, denoted by \(\bbz_t = \ccalE_{\theta}(\bbA_t, \bbX_t)\), where \(\bbz_t \in \mathbb{R}^{1 \times F}\).
Hence, the objective is to learn \emph{discriminative representations for entire graphs rather than individual nodes}.  
Similar to node-level approaches, InfoNCE-based methods are widely adopted in this setting~\cite{graphcl, ad-gcl, joao}.
These methods generate two augmented views of the input graph through transformations such as edge perturbation, feature masking, or subgraph sampling.  
Following this, let \(\bbz_i\) and \(\tbz_i\) denote the graph-level representations of the original and augmented views, respectively.  
The InfoNCE loss is then used to bring \(\bbz_i\) and \(\tbz_i\) (positive pair) closer in the embedding space while pushing them apart from representations of all other graphs in the batch (\(\tbz_j\), for all \(j \neq i\), i.e., negative samples).  

\subsection{Graphon} 
\label{subsec:background_graphon}

A graphon is defined as a bounded, symmetric, and measurable function $\ccalW : [0,1]^2 \rightarrow [0,1]$~\citep{graphon}.
By construction, a graphon acts as a \emph{generative model for random graphs}, allowing the sampling of graphs that exhibit similar structural properties.
To generate an undirected graph $\ccalG$ with $N$ nodes from a given graphon $\ccalW$, the process consists of two main steps: (1) assigning each node a latent variable drawn uniformly at random from the interval $[0,1]$, and (2) connecting each pair of nodes with a probability given by evaluating $\ccalW$ at their respective latent variable values.
Formally, the steps are as follows:
\begin{align}
\label{eq:stochastic_sampling}
    \bbeta(i) &\sim \operatorname{Uniform} ([0,1]),\quad \forall \; i=1,\cdots,N, \\ \nonumber
    \bbA(i,j) &\sim \text{Bernoulli}\left(\ccalW(\bbeta(i), \bbeta(j))\right),\quad \forall\; i,j=1,\cdots,N, 
\end{align}
where the latent variables $\bbeta(i) \in [0,1]$ are independently drawn for each node $i$.

\paragraph{Graphon estimation.}
The generative process in~\eqref{eq:stochastic_sampling} can also be viewed in reverse: given a collection of graphs (represented by their adjacency matrix) \(\ccalD = \{\bbA_t\}_{t=1}^{M}\) that are sampled from an \emph{unknown} graphon \(\ccalW\), estimate \(\ccalW\).
Several methods have been proposed for this task~\citep{sas, sassbm, gwb, ignr, sigl}.
We focus on SIGL~\citep{sigl}, a resolution-free method that, in addition to estimating the graphon, also \emph{infers the latent variables $\bbeta$}, making it particularly useful for model-driven augmentation in GCL.
This method parameterizes the graphon using an implicit neural representation (INR) \citep{siren}, a neural architecture defined as \(f_{\phi}(x, y): [0,1]^2 \rightarrow [0,1]\) where the inputs are coordinates from $[0,1]^2$ and the output approximates the graphon value $\ccalW$ at a particular position. 
In a nutshell, SIGL works in three steps: (1) a sorting step using a GNN \(g_{\phi'}(\bbA)\) that estimates the latent node positions or representations $\bbeta$; 
(2) a histogram approximation of the sorted adjacency matrices; 
and (3) learning the parameters $\phi$ by minimizing the mean squared error between $f_{\phi}(x, y)$ and the histograms (obtained in step 2). 
More details of SIGL are provided in Appendix~\ref{app:graphon}.



\section{Model-driven graph contrastive learning (MGCL)}
\label{sec:MGCL}

MGCL estimates an underlying generative model (graphons in this work) for the data and leverages it to guide the GCL pipeline. 
We first describe the MGCL framework for node-level tasks in Section~\ref{subsec:methodNode}, and then extend it to graph-level tasks in Section~\ref{subsec:methodGraph}. 
A detailed algorithm of MGCL is included in Appendix~\ref{app:alg}.

\subsection{Node-level MGCL}
\label{subsec:methodNode}

In node-level tasks, such as node classification, the input is a single, and usually large, graph $\ccalG = (\ccalV, \bbA, \bbX)$ with $|\ccalV| = N$ nodes. 
As detailed in Section~\ref{subsec:background_gcl}, the goal is to train an encoder \(\bbH = \ccalE_{\theta}(\bbA, \bbX)\) in a self-supervised manner, so that the node embeddings $\bbH$ can be used for downstream node-level tasks.
We assume that the \emph{observed graph \(\ccalG\) is sampled from an underlying single graphon model \(\ccalW\)}.
If we have access to $\ccalW$, we can use it to generate more informative positive and negative views for contrastive learning. 
However, the graphon model is generally unknown.

\paragraph{Graphon-informed augmentation.}
As a first step, we estimate the underlying graphon \(\ccalW\) and the sampling process that generates \(\ccalG\) from it.
To this end, we adopt the method proposed in SIGL~\cite{sigl}.
As described in Section~\ref{subsec:background_graphon}, the outputs of SIGL are two models: $(i)$ a learned graphon INR \(f_{\phi}(x, y): [0,1]^2 \rightarrow [0,1]\), which is an estimator of \(\ccalW\); and $(ii)$ a learned GNN \(g_{\phi'}(\bbA)\), which produces node-wise latent representations \(\bbeta\) aligned with the graphon domain.
These outputs enable a graphon-informed augmentation (GIA) procedure formally described as:
\begin{equation}
\tbA \sim \ccalT_{\text{GIA}}(\bbA, f_{\phi}, g_{\phi'}),
\end{equation}
where \(\bbA\) and \(\tbA\) denote the original and augmented adjacency matrices, respectively.

The transformation \(\ccalT_{\text{GIA}}\) proceeds as follows:  
(1) we first randomly select a subset \(E_{\text{sel}}~\subset~\{(i,j) \mid 1 \leq i < j \leq n\}\) containing \(r\%\) of all node pairs, such that \(|E_{\text{sel}}| = (r/100)~\times~\left({n(n-1)}/{2}\right)\);  
(2) for each selected pair \((i, j) \in E_{\text{sel}}\), we resample the corresponding adjacency entry using the probability assigned by the learned graphon INR:
\begin{equation}
\tbA(i,j) \sim \text{Bernoulli}\left(f_{\phi}(\bbeta(i), \bbeta(j))\right), \quad \tbA(j,i) = \tbA(i,j);
\end{equation}
where \(\bbeta = g_{\phi'}(\bbA)\) and (3) for all remaining pairs \((i,j) \notin E_{\text{sel}}\), we retain the original entries: \(\tbA(i,j) = \bbA(i,j)\).
This transformation introduces structure-aware perturbations informed by the graphon generative process, resulting in more meaningful augmentations compared to random edge modifications. 
{Although the transformation \(\ccalT_{\text{GIA}}\) follows a fixed procedure, it is inherently \textit{stochastic} due to the random selection of node pairs \(E_{\text{sel}}\) and the Bernoulli resampling of edge entries. 
As a result, it defines a distribution over possible augmentations conditioned on the graphon and the input graph, from which each augmented view is sampled.}
In fact, standard random edge perturbation can be viewed as a special case of GIA, where the graphon is assumed to be a constant function \(\ccalW = 0.5\), and each edge is resampled with uniform probability $0.5$. 
In contrast, learning the graphon from the observed graph enables GIA to assign edge-specific probabilities based on the estimated generative structure.

\paragraph{Encoder training.}
To train the encoder \(\ccalE_{\theta}\), we adopt the DGI~\cite{dgi} objective to maximize mutual information between local (node-level) and global (graph-level) representations. 
Specifically, using the DGI loss, we seek to encourage node embeddings to be predictive of a global summary vector derived from the positive view, while discouraging alignment with representations from a corrupted version of the graph (negative view).

In our setup, the positive view is generated using the GIA procedure, where the augmented adjacency matrix \(\tbA_p \sim \ccalT_{\text{GIA}}(\bbA, f_{\phi}, g_{\phi'})\) is obtained using the transformation \(\ccalT_{\text{GIA}}\), resulting in \(\ccalG_p = (\ccalV, \tbA_p, \bbX)\). 
The negative view \(\ccalG_n = (\ccalV, \tbA_n, \tbX)\) is constructed similarly via $\ccalT_{\text{GIA}}$, with an additional feature shuffling applied to the node attributes $\bbX$ to generate a corrupted view $\tbX$.
The encoder \(\ccalE_{\theta}\) produces node representations \(\bbH = \ccalE_{\theta}(\bbA, \bbX)\) and \({\tbH_n} = \ccalE_{\theta}(\tbA_n, \tbX)\) for the original and negative graphs, respectively. 
Similar to GraphCL~\cite{graphcl}, a summary vector \(\bbs\) is computed by applying a readout function to the node embeddings of the positive augmented graph \(\tbH_p = \ccalE_{\theta}(\tbA_p, \bbX)\), capturing the global graph-level context as \(\bbs_p = \ccalR(\tbH_p)\).

A discriminator \(\ccalD(\cdot, \cdot)\) then scores the agreement between node embeddings and the summary vector. 
The DGI loss is defined as:
\begin{equation}
\label{eq:dgi}
\ccalL_{\text{DGI}} = \frac{1}{2N} \left( \sum_{i=1}^{N} \log \ccalD(\bbh_i, \bbs_p) + \log\left(1 - \ccalD(\tbh_i^n, \bbs_p)\right) \right),
\end{equation}
where \(\bbh_i = [\bbH]_{i,:}\) and \(\tbh_i^n=[\tbH_n]_{i,:}\) are the representations of node \(i\) from the original and negative views, respectively.
Minimizing~\eqref{eq:dgi} promotes alignment between each node and the overall graph semantics in the positive view, while forcing representations from the negative view to be dissimilar.
Notably, this loss is equivalent to a binary cross-entropy (BCE) loss, where positive samples \((\bbh_i, \bbs_p)\) are labeled as 1 and negative samples \((\tbh_i^n, \bbs_p)\) as 0.

A key distinction in our framework lies in how the positive and negative views are generated. 
Unlike standard GCL methods that rely on heuristic augmentations (e.g, random edge dropping), potentially disrupting meaningful structural patterns, our approach leverages an estimated graphon model to guide this process in a principled way. 
By treating the \emph{graphon as a generative prior}, MGCL introduces perturbations that better reflect the underlying distribution of the data.
Consequently, the summary vector \(\bbs_p\) can be interpreted as the \emph{representation of a related sample drawn from the same generative process}, making it a more semantically meaningful anchor for the DGI loss in~\eqref{eq:dgi}.

\subsection{Graph-level MGCL}
\label{subsec:methodGraph}

In graph-level tasks, such as graph classification, we are no longer dealing with a single graph. 
Instead, we are given a set of graphs \(\{\ccalG_t\} = \{(\ccalV_t, \bbA_t, \bbX_t)\}_{t=1}^L\), and the goal is to train an encoder that generates a representation for each graph, denoted by \(\bbz_t = \ccalE_{\theta}(\bbA_t, \bbX_t)\), in a self-supervised manner.
Similar to the node-level setting, we seek to guide contrastive learning using a graphon generative model. 
Unlike node-level tasks, however, it is unreasonable to assume that all graphs $\ccalG_t$ are samples from a single graphon; therefore, we model the dataset as being generated from multiple graphons.

\paragraph{Graph clustering.}
We begin by clustering the observed graphs so that those within the same cluster are more likely to have been generated by the same graphon.
To perform this clustering, we pass each graph through a randomly initialized GNN \(g_{\zeta}(\cdot)\) to obtain initial graph-level representations \( \bbz_t^{\text{init}} = g_{\zeta}(\bbA_t, \bbX_t) \).
Since graphs sampled from the same graphon tend to share structural similarities, they will exhibit similar embeddings under a shared GNN~\cite{graphonwnn}. 
We then apply \(k\)-means clustering to the embeddings \(\{\bbz_t^{\text{init}}\}_{t=1}^L\), partitioning the graphs into \(K=\log(L)\) clusters. 
This yields clusters of graphs with similar representations, suggesting that the same underlying graphon likely generates graphs within each cluster.

\paragraph{Multiple models estimation.}
Next, we estimate $K$ graphons \(\{\ccalW^{(k)}\}_{k=1}^K\) with SIGL for each cluster based only on the graphs assigned to each cluster.
This results in a set of model pairs \(\{(f_{\phi}^{(k)}, g_{\phi'}^{(k)})\}_{k=1}^K\), where each \(f_{\phi}^{(k)}\) serves as a graphon estimator for cluster \(k\), and each \(g_{\phi'}^{(k)}\) produces the node-wise latent variables for graphs assigned to the $k$-cluster.
Each \(\ccalW^{(k)}\) serves as the \textit{positive} generative model for graphs in cluster \(k\), and as a \textit{negative} model for graphs in all other clusters.

Having multiple models allows us to learn graph representations such that each graph is encouraged to align closely with its corresponding positive model while remaining dissimilar to the negative models associated with other clusters. 
This contrasts with conventional approaches that operate at the individual graph level, where each graph is pushed away from all others while being pulled toward a randomly augmented version of itself.
To incorporate the model information, we consider two strategies: $(i)$ augment each graph using only its positive model to preserve the structure of similar graphs within its cluster, and $(ii)$ contrast it specifically against (augmented views of) graphs generated from different graphons belonging to other clusters.

\paragraph{Graphon-informed augmentation.}
Specifically, given the original adjacency matrix \(\bbA_t\) and its cluster \(c(t)\), its assigned estimated graphon \(f_{\phi}^{c(t)}(\cdot)\), and the latent variables \(\bbeta_t = g_{\phi'}^{c(t)}(\bbA_t)\) of the nodes on the graph, two distinct augmented adjacency matrices are obtained as:
\begin{equation}
\tbA_t^1, \tbA_t^2 \sim \ccalT_{\text{GIA}}(\bbA_t, f_{\phi}^{c(t)}, g_{\phi'}^{c(t)}), 
\end{equation}
%
As explained in the node case, \(\tbA_t^1\) and \(\tbA_t^2\) represent two distinct augmented views, resulting from the \textit{stochastic} behavior of \(\ccalT_{\mathrm{GIA}}\).
Consequently, for each graph \(t\), the encoder produces a representation \(\bbz_t^1 = \ccalE_{\theta}(\tbA_t^1, \bbX_t)\) for the first view, and \(\bbz_t^2 = \ccalE_{\theta}(\tbA_t^2, \bbX_t)\) for the second view.

\paragraph{Model-aware contrasting.}
To train the encoder \(\ccalE_{\theta}\), we adopt a modified InfoNCE loss defined as:
\begin{equation}
\label{eq:loss_cluster}
\ell_t = -\log \frac{\exp (\text{sim}(\bbz_t, \bbz_t^1) / \tau)}{\sum_{t'=1,\, c(t') \neq c(t)}^L \exp (\text{sim}(\bbz_t, \bbz_{t'}^2) / \tau)}, \quad \ccalL_{\text{all}} = \frac{1}{L}\sum_{t=1}^{L}\ell_t.
\end{equation}
where \(\text{sim}(\cdot, \cdot)\) denotes a similarity function (e.g., cosine similarity), \(\tau\) is a temperature parameter, \(c(t)\) denotes the cluster assignment of graph \(t\), \(L\) is the number of graphs in the batch, and \(\bbz_t\), \(\bbz_t^1\), and \(\bbz_{t'}^2\) are the representations of the original graph, its first augmented view, and second augmented view of other graphs respectively.

Minimizing the loss in~\eqref{eq:loss_cluster} encourages each graph to be similar to its graphon-informed first view while being dissimilar to the second views of graphs that belong to different clusters. 
As a result, unlike traditional InfoNCE-based methods that push each graph away from all other graphs~\cite{graphcl}, our formulation only contrasts a graph against structurally dissimilar graphs modeled by different graphons. 
By doing so, the model avoids penalizing similar graphs and focuses on distinguishing between different generative patterns.

Note that if we were to generate only a single augmentation for all graphs, we would be pushing \(\bbz_t\) to be similar to \(\bbz_t^1\), while simultaneously pushing all other graph representations \(\bbz_{t',t' \neq t}\), away from this same augmentation \(\bbz_t^1\). 
This setup encourages dissimilarity among all graphs, mirroring existing contrastive approaches.
However, this contradicts the core idea of our method, which is to separate graphs based on their underlying generative models rather than on an individual basis. 

Moreover, the estimated graphon in each cluster is learned from multiple graphs, capturing the structure of an individual graph and the shared patterns among structurally similar graphs. 
This makes the graphon-informed augmentation more meaningful, incorporating information derived from a richer, cluster-level understanding of graph structure.

\section{Experiments}
\label{sec:numerical}

In this section, we evaluate MGCL through comparisons with a variety of baseline methods across both node-level tasks (node classification and clustering) and graph-level tasks (graph classification). 
Additionally, we conduct additional experiments to gain deeper insights into the behavior and effectiveness of MGCL. 
A description of the datasets, baselines, experimental setups, and hyperparameters is provided in Appendix~\ref{app:implementation}, while additional ablation studies are presented in Appendix~\ref{app:experiments}.

\paragraph{Setup.} 
For node-level tasks, we use six widely adopted networks: Cora~\cite{sen2008collective}, CiteSeer~\cite{sen2008collective}, PubMed~\cite{sen2008collective}, Photo~\cite{shchur2018pitfalls}, Cs~\cite{shchur2018pitfalls}, and Physics~\cite{shchur2018pitfalls}. 
Baseline models include: a supervised graph neural network (GCN)~\cite{gcn}, a classical graph representation method (Node2vec~\cite{node2vec}), and seven self-supervised graph contrastive learning methods—DGI~\cite{dgi}, MVGRL~\cite{mvgrl}, GRACE~\cite{grace}, GCA~\cite{gca}, GraphCL~\cite{graphcl}, BGRL~\cite{bgrl}, and GBT~\cite{gbt}.

For graph-level tasks, we evaluate on eight datasets from the TUDataset~\cite{TUDataset} collection: IMDB-B, RDT-B, RDT-M5K, COLLAB, MUTAG, DD, PROTEINS, and NCI1. 
Baselines include graph kernel-based methods (graphlet kernel (GL), Weisfeiler-Lehman sub-tree kernel (WL), deep graph kernel (DGK)), three unsupervised graph-level representation learning methods (Node2vec~\cite{node2vec}, Sub2vec~\cite{sub2vec}, Graph2vec~\cite{graph2vec}), and four state-of-the-art GCL-based methods (MVGRL~\cite{mvgrl}, InfoGraph~\cite{infograph}, GraphCL~\cite{graphcl}, and JOAO~\cite{joao}).

\paragraph{Evaluation.} 

The experiments follow the linear evaluation protocol~\cite{peng2020graph}, where models are first trained in an unsupervised manner, and the resulting embeddings are subsequently used for downstream tasks. 
The graph encoder \(\ccalE_\theta(\cdot)\) is implemented as a two-layer GCN~\cite{gcn} for node-level tasks and as a three-layer GIN~\cite{gin} for graph classification tasks.
For node classification, each graph is split into training, validation, and test sets in a 1:1:8 ratio. 
A single-layer linear classifier is trained on the training set and evaluated on the test set. 
This process is repeated 10 times with different random splits, and we report the mean accuracy along with the standard deviation.
For node clustering, we apply K-means to the learned node embeddings and compare the resulting clusters to the ground truth labels. 
For graph classification, we adopt the evaluation protocol from~\cite{infograph}, performing 10-fold cross-validation on each dataset. 
The resulting graph-level embeddings are used to train an SVM classifier, and we report the average performance across the folds.
To summarize performance, we assign a rank to each method based on its performance on each dataset, separately for the node and graph classification tasks. 
The Average Rank (A.R.) is then computed as the mean of these ranks across all datasets.

\subsection{Node-level experimental results}
\label{result_real}

\paragraph{Node classification.}
\begin{table*}[t]
\begin{center}
\caption{Accuracy in percentage (mean$\pm$std) over ten trials of node classification. 
\small{The best and two runner-up methods are highlighted in \textbf{bolded} and \underline{underlined}, respectively.}}
\label{tab:node_classification}
\resizebox{0.90\textwidth}{!}{
\begin{tabular}{l|l|cccccc|c}
\toprule
Model & Input & Cora & CiteSeer & PubMed & Photo & Cs & Physics & A.R. \\
\midrule
Raw Features & X & 53.30{\scriptsize$\pm$1.99} & 57.40{\scriptsize$\pm$0.77} & 79.97{\scriptsize$\pm$0.45} & 79.99{\scriptsize$\pm$0.90} & 89.76{\scriptsize$\pm$0.30} & 94.32{\scriptsize$\pm$0.14} & 10.33  \\
Node2vec & A & 77.06{\scriptsize$\pm$0.77} & 53.39{\scriptsize$\pm$1.56} & 79.40{\scriptsize$\pm$0.36} & 89.67{\scriptsize$\pm$0.12} & 85.08{\scriptsize$\pm$0.03} & 91.19{\scriptsize$\pm$0.04} & 10.67  \\
GCN & A, X, Y & 82.32{\scriptsize$\pm$1.79} & 72.13{\scriptsize$\pm$1.17} & 84.90{\scriptsize$\pm$0.38} & 92.42{\scriptsize$\pm$0.22} & 93.03{\scriptsize$\pm$0.31} & 95.65{\scriptsize$\pm$0.16} & 6.17  \\
\midrule
DGI & A, X & 82.60{\scriptsize$\pm$0.40} & 71.49{\scriptsize$\pm$0.14} & \underline{86.00}{\scriptsize$\pm$0.14} & 91.49{\scriptsize$\pm$0.25} & 92.15{\scriptsize$\pm$0.63} & 94.51{\scriptsize$\pm$0.52} & 7.5 \\
MVGRL & A, X & 83.03{\scriptsize$\pm$0.27} & 72.75{\scriptsize$\pm$0.46} & {85.63}{\scriptsize$\pm$0.38} & 92.01{\scriptsize$\pm$0.13} & 92.11{\scriptsize$\pm$0.12} & 95.33{\scriptsize$\pm$0.03} & 6.33 \\
GRACE & A, X & 83.30{\scriptsize$\pm$0.40} & 71.41{\scriptsize$\pm$0.38} & \textbf{86.51}{\scriptsize$\pm$0.39} & 92.65{\scriptsize$\pm$0.28} & 92.17{\scriptsize$\pm$0.04} & 95.26{\scriptsize$\pm$0.06} & 5.67\\
GCA & A, X & \underline{83.90}{\scriptsize$\pm$0.41} & 72.21{\scriptsize$\pm$0.46} & \underline{86.01}{\scriptsize$\pm$0.75} & \underline{92.78}{\scriptsize$\pm$0.17} & 93.10{\scriptsize$\pm$0.01} & \underline{95.68}{\scriptsize$\pm$0.05} & \underline{3.17} \\
GraphCL & A, X & 83.64{\scriptsize$\pm$0.54} & \underline{72.37}{\scriptsize$\pm$0.67} & 85.61{\scriptsize$\pm$0.13} & 92.22{\scriptsize$\pm$0.42} & \underline{93.12}{\scriptsize$\pm$0.20} & \textbf{95.80}{\scriptsize$\pm$0.14} & 4.50 \\
BGRL & A, X & 83.77{\scriptsize$\pm$0.75} & 71.99{\scriptsize$\pm$0.42} & 84.94{\scriptsize$\pm$0.17} & \textbf{93.24}{\scriptsize$\pm$0.29} & \underline{93.31}{\scriptsize$\pm$0.13} & 95.63{\scriptsize$\pm$0.04} & \underline{4.17}\\
GBT & A, X & \underline{83.89}{\scriptsize$\pm$0.66} & \underline{72.57}{\scriptsize$\pm$0.61} & 85.71{\scriptsize$\pm$0.32} & 92.63{\scriptsize$\pm$0.44} & 92.95{\scriptsize$\pm$0.17} & 95.07{\scriptsize$\pm$0.17} & 5.00 \\
\midrule
MGCL & A, X & \textbf{84.08}{\scriptsize$\pm$0.67} & \textbf{73.26}{\scriptsize$\pm$0.55} & 85.72{\scriptsize$\pm$0.25} & \underline{92.78}{\scriptsize$\pm$0.23} & \textbf{93.42}{\scriptsize$\pm$0.10} & \textbf{95.80}{\scriptsize$\pm$0.08} & \textbf{1.67} \\
\bottomrule
\end{tabular}}
\end{center}
\vspace{-0.3cm}
\end{table*}

The results for this task, shown in Table~\ref{tab:node_classification}, demonstrate the effectiveness of the proposed MGCL framework. 
MGCL achieves the best performance on four out of six datasets (Cora, CiteSeer, Cs, and Physics) and the second-best on Photo.
It consistently outperforms both classical and supervised baselines across all datasets. 
These improvements are especially notable when compared with methods like GraphCL, GRACE, GCA, or MVGRL, which rely on random or hand-crafted augmentation strategies. 
MGCL’s superior performance can be attributed to its use of graphon-informed augmentations, which introduce structure-aware perturbations tailored to the underlying graph distribution. 
Furthermore, MGCL achieves the lowest average rank (A.R.) of 1.67, significantly outperforming all other methods, highlighting its consistent advantage across diverse datasets.

\paragraph{Node clustering.}
\begin{wraptable}{r}{0.50\textwidth}
\vspace{-5mm}
\centering
\caption{\small{Node clustering performance: NMI \& ARI Scores in percentage (mean$\pm$std).}}
\label{tab:node_clustering}
\resizebox{0.49\textwidth}{!}{
\begin{tabular}{l|cc|cc}
\toprule
& \multicolumn{2}{c|}{Cora} & \multicolumn{2}{c}{CiteSeer}\\
& NMI & ARI & NMI & ARI\\
\midrule
DGI & 52.75{\scriptsize$\pm$0.94} & 47.78{\scriptsize$\pm$0.65} & 40.43{\scriptsize$\pm$0.81} & 41.84{\scriptsize$\pm$0.62} \\
MVGRL & 54.21{\scriptsize$\pm$0.25} & 49.04{\scriptsize$\pm$0.67} & 43.26{\scriptsize$\pm$0.48} & 42.73{\scriptsize$\pm$0.93}\\
GRACE & 54.59{\scriptsize$\pm$0.32} & 48.31{\scriptsize$\pm$0.63} & 43.02{\scriptsize$\pm$0.43} & 42.32{\scriptsize$\pm$0.81}\\
GBT & 55.32{\scriptsize$\pm$0.65} & 48.91{\scriptsize$\pm$0.73} & 44.01{\scriptsize$\pm$0.97} & 42.61{\scriptsize$\pm$0.63}\\
\midrule
MGCL & \textbf{56.56}{\scriptsize$\pm$0.67} & \textbf{52.19}{\scriptsize$\pm$1.20} & \textbf{44.07}{\scriptsize$\pm$0.43} & \textbf{44.22}{\scriptsize$\pm$1.02} \\
\bottomrule
\end{tabular}}
\vspace{-3mm}
\end{wraptable}
The results are reported in Table~\ref{tab:node_clustering}.
We observe that MGCL consistently outperforms all baselines across both datasets—Cora and CiteSeer—in terms of both NMI and ARI scores. 
Specifically, MGCL achieves the highest NMI and ARI on Cora with 56.56\% and 52.19\%, respectively, improving upon the best-performing baseline by margins of 1.24 (NMI) and 3.15 (ARI).
On CiteSeer, MGCL also achieves the best results with 44.07\% NMI and 44.22\% ARI, demonstrating a clear advantage over the second-best method, MVGRL. 
These results highlight the effectiveness of MGCL in capturing meaningful and discriminative node-level representations that are better-aligned with true labels using the underlying model.

\paragraph{Effect of Graphon-Informed Augmentation (GIA).}

To evaluate the impact of GIA and the incorporation of the underlying graphon model in the augmentation process, we compare MGCL with a variant of GraphCL~\cite{graphcl}, referred to as GraphCL-EP.
In this variant, augmentations are generated using random edge perturbations—edges are randomly dropped or added—rather than being guided by the graphon.
This experiment is conducted in a synthetic setting for a node classification task. 

We construct a synthetic graph by sampling from a ground-truth graphon \(\ccalW\) using the procedure described in Section~\ref{subsec:background_graphon}. 
Each node is assigned a latent variable \(\bbeta(i) \sim \text{Uniform}[0,1]\), and node features are generated by applying a nonlinear transformation: \(\bbx_i = f(\eta_i)\). 
To create the labels, we pass the graph \(\ccalG\) and \(\ccalW\) through a graphon neural network (WNN)~\cite{graphonwnn} that uses the graph structure and latent variables to produce the ground-truth node labels.
The goal is to perform node classification on this synthetic dataset using the evaluation procedure described earlier in Section~\ref{result_real}. 
We compare four methods: (i) a logistic regression (LR) model trained directly on the raw features; (ii) a semi-supervised GCN; (iii) MGCL; and (iv) GraphCL-EP.

\begin{table*}[h]
\vspace{-3mm}
\begin{center}
\caption{Node classification results on synthetic data.}
\label{tab:node_sim}
\small
\resizebox{0.60\textwidth}{!}{
\begin{tabular}{c|c|c|c|c}
 & Raw Features & GCN & MGCL & GraphCL-EP \\
\hline
Acc(\%) & 99.40{\scriptsize$\pm$0.15} & 90.56{\scriptsize$\pm$2.04} & 86.38{\scriptsize$\pm$2.46} & 80.81{\scriptsize$\pm$1.57}  \\
\end{tabular}}
\end{center}
\vspace{-5mm}
\end{table*}

As shown in Table~\ref{tab:node_sim}, the LR model trained on raw features achieves nearly perfect accuracy, which is expected since both the features and labels are generated from the same latent variables. 
Notably, the performance of GraphCL-EP is significantly lower than that of MGCL, with a 5.57\% drop in accuracy.  
Moreover, MGCL (86.38\%) performs much closer to the supervised GCN (90.56\%) than GraphCL-EP does.  
This highlights the effectiveness of leveraging an estimated graphon model to guide structural perturbations, rather than relying on random perturbations.


\subsection{Graph-level experimental results}
\label{}
\paragraph{Graph classification.}
As shown in Table~\ref{tab:unsupervised}, MGCL achieves strong performance across a wide range of graph-level benchmarks from the TUDataset collection~\cite{TUDataset}. 
MGCL ranks first on five out of eight datasets and is among the top 3 methods in the other 3 datasets. 
On the NCI1 dataset, MGCL achieves the best performance among all GCL-based methods. 
For the IMDB-B dataset, InfoGraph performs best, suggesting that augmentation-based strategies may be less effective for this particular benchmark. 
Nonetheless, MGCL outperforms other augmentation-based methods such as GraphCL and JOAO.
Most notably, MGCL obtains the lowest average rank (A.R.) of 1.75, outperforming all competing baselines, including those specifically designed for graph-level tasks. 
These results highlight MGCL’s ability to learn generalizable and transferable graph-level representations. 

\begin{table*}[t]
\begin{center}
\caption{Unsupervised representation learning on TUDataset.
\small{
The best and two runner-up methods are highlighted in \textbf{bolded} and \underline{underlined}, respectively.
Results are taken from~\cite{joao}.}
}
\label{tab:unsupervised}
\resizebox{\textwidth}{!}{
\begin{tabular}{c|c c c c|c c c c|c}
    \hline
    Methods & NCI1 & PROTEINS & DD & MUTAG & COLLAB & RDT-B & RDT-M5K & IMDB-B & A.R. \\
    \hline
    GL & - & - & - & 81.66{\scriptsize$\pm$2.11} & - & 77.34{\scriptsize$\pm$0.18} & 41.01{\scriptsize$\pm$0.17} & 65.87{\scriptsize$\pm$0.98} & 7.5 \\
    WL &  \underline{80.01}{\scriptsize$\pm$0.50} & 72.92{\scriptsize$\pm$0.56} & - & 80.72{\scriptsize$\pm$3.00} & - & 68.82{\scriptsize$\pm$0.41} & 46.06{\scriptsize$\pm$0.21} &  \underline{72.30}{\scriptsize$\pm$3.44} & 5.8 \\
    DGK &  \textbf{80.31}{\scriptsize$\pm$0.46} & 73.30{\scriptsize$\pm$0.82} & - & 87.44{\scriptsize$\pm$2.72} & - & 78.04{\scriptsize$\pm$0.39} & 41.27{\scriptsize$\pm$0.18} & 66.96{\scriptsize$\pm$0.56} & 4.8 \\
    node2vec & 54.89{\scriptsize$\pm$1.61} & 57.49{\scriptsize$\pm$3.57} & - & 72.63{\scriptsize$\pm$10.20} & - & - & - & - & 8.7 \\
    sub2vec & 52.84{\scriptsize$\pm$1.47} & 53.03{\scriptsize$\pm$5.55} & - & 61.05{\scriptsize$\pm$15.80} & - & 71.48{\scriptsize$\pm$0.41} & 36.68{\scriptsize$\pm$0.42} & 55.26{\scriptsize$\pm$1.54} & 9.5 \\
    graph2vec & 73.22{\scriptsize$\pm$1.81} & 73.30{\scriptsize$\pm$2.05} & - & 83.15{\scriptsize$\pm$9.25} & - & 75.78{\scriptsize$\pm$1.03} & 47.86{\scriptsize$\pm$0.26} & 71.10{\scriptsize$\pm$0.54} & 6.0 \\
    \hline
    MVGRL & - & - & - & 75.40{\scriptsize$\pm$7.80} & - & 82.00{\scriptsize$\pm$1.10} & - & 63.60{\scriptsize$\pm$4.20} & 7.7 \\
    InfoGraph & 76.20{\scriptsize$\pm$1.06} &  \underline{74.44}{\scriptsize$\pm$0.31} & 72.85{\scriptsize$\pm$1.78} &  \underline{89.01}{\scriptsize$\pm$1.13} &  \underline{70.65}{\scriptsize$\pm$1.13} & 82.50{\scriptsize$\pm$1.42} & 53.46{\scriptsize$\pm$1.03} &  \textbf{73.03}{\scriptsize$\pm$0.87} & \underline{3.4} \\
    GraphCL & 77.87{\scriptsize$\pm$0.41} &  {74.39}{\scriptsize$\pm$0.45} &  \underline{78.62}{\scriptsize$\pm$0.40} & 86.80{\scriptsize$\pm$1.34} &  \underline{71.36}{\scriptsize$\pm$1.15} &  \underline{89.53}{\scriptsize$\pm$0.84} &  \textbf{55.99}{\scriptsize$\pm$0.28} &  {71.14}{\scriptsize$\pm$0.44} &  \underline{3.1} \\
    JOAO &  {78.07}{\scriptsize$\pm$0.47} &  \underline{74.55}{\scriptsize$\pm$0.41} &  \underline{77.32}{\scriptsize$\pm$0.54} &  \underline{87.35}{\scriptsize$\pm$1.02} &  {69.50}{\scriptsize$\pm$0.36} &  \underline{85.29}{\scriptsize$\pm$1.35} &  \underline{55.74}{\scriptsize$\pm$0.63} & 70.21{\scriptsize$\pm$3.08} & 3.5 \\
    \hline
    MGCL & \underline{78.66}{\scriptsize$\pm$0.34} & \textbf{74.85}{\scriptsize$\pm$0.71} &  \textbf{78.88}{\scriptsize$\pm$0.38} &  \textbf{89.55}{\scriptsize$\pm$1.08} & \textbf{71.44}{\scriptsize$\pm$0.71} &  \textbf{90.25}{\scriptsize$\pm$0.39} &  \underline{55.65}{\scriptsize$\pm$0.32} & \underline{71.55}{\scriptsize$\pm$0.53} &  \textbf{1.75} \\
    \hline
\end{tabular}}
\end{center}
\end{table*}


\paragraph{Effect of model-aware clustering on false negative reduction.}

\begin{wrapfigure}{r}{0.55\textwidth}  
    \vspace{-3mm}
    \centering
    \includegraphics[width=0.54\textwidth]{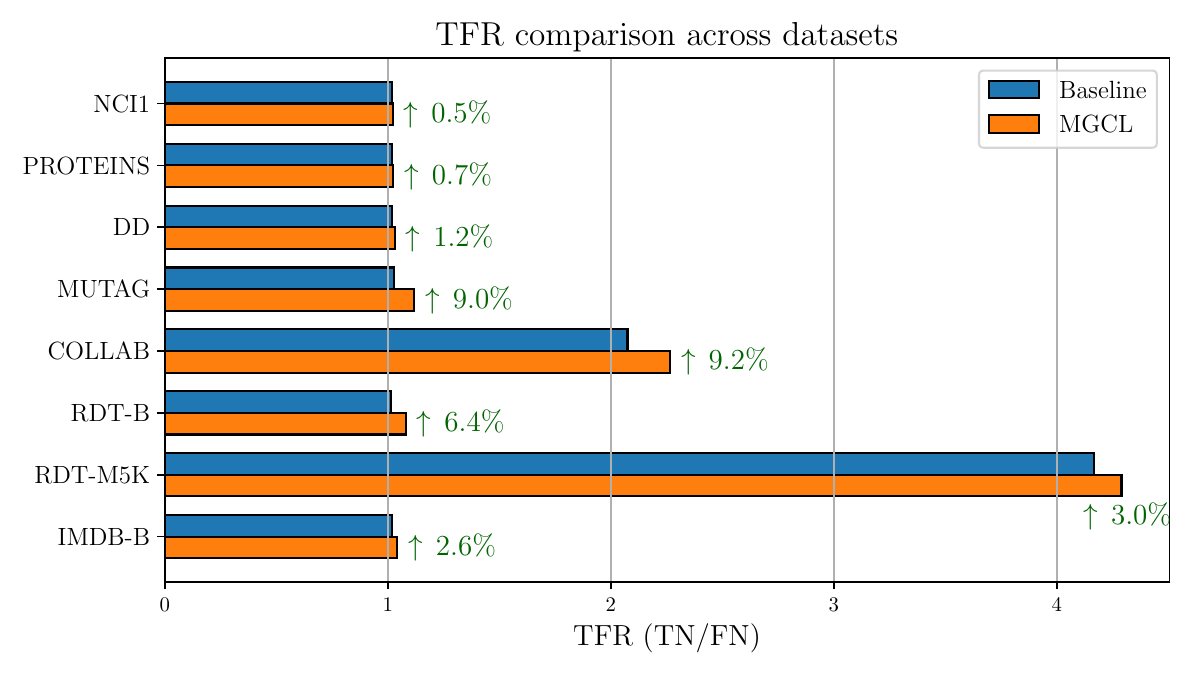}
    \caption{\small{Effect of clustering on TFR across different datasets.}}
    \label{fig:tfr}
    \vspace{-3mm}
\end{wrapfigure}

To evaluate the effect of clustering on the rate of false negatives, we define the True Negative to False Negative Ratio (TFR).
To compute this metric, in each data batch, we examine the negative samples relative to a graph \(i\).
Among these, the negative samples that share the same class as graph \(i\) are considered false negatives, while those with a different class are treated as true negatives.

Note that in InfoNCE-based methods, all graphs in the batch except graph \(i\) itself are treated as negative samples.
However, in MGCL, the set of negative samples is smaller, as we exclude graphs from the same cluster as \(i\).
Although this reduced set naturally results in fewer false negatives, computing the relative ratio of true negatives to false negatives (TFR) ensures a fair comparison across methods.
We compute the TFR for each graph in the batch and then average it across all graphs in the dataset.
As shown in Figure~\ref{fig:tfr}, this metric increases across all datasets compared to the baseline (which represents all InfoNCE-based methods).
Notably, the increase is more significant in the MUTAG, COLLAB, and RDT-B datasets, where MGCL also outperforms existing methods, as reported in Table~\ref{tab:unsupervised}.

\paragraph{Underlying models.}

\begin{figure*}[t]
	\centering
	\includegraphics[width=\textwidth]{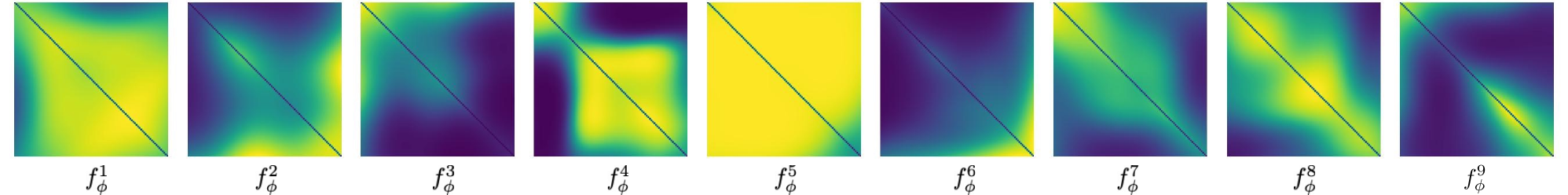}
	\caption{\small{Cluster-specific estimated graphons in the COLLAB dataset,                revealing diverse structures.}}
	\label{fig:graphons}
\end{figure*}

In Figure~\ref{fig:graphons}, we present the estimated graphons for the COLLAB dataset, each corresponding to a cluster identified by our framework.
The estimated graphons display diverse structural patterns: Cluster 4 exhibits a block-like structure similar to a two-community stochastic block model (SBM) with an imbalanced size ratio, Clusters 1 and 5 display nearly uniform connectivity, suggesting dense or complete graph structures, and Cluster 6 reveals a heavy-tailed pattern indicative of power-law behavior. 
Although some similarities exist among certain models -- likely due to the relatively high number of estimated graphons -- the overall variability emphasizes the presence of multiple distinct generative mechanisms within the dataset.
This heterogeneity demonstrates the limitations of using a single fixed or random augmentation strategy, as commonly adopted in existing GCL methods.
Furthermore, the presence of such diverse models reinforces the need for model-aware contrastive learning to reduce false negatives.
For instance, it is unreasonable to generate highly dissimilar representations for graphs in Cluster 5, where most graphs are nearly complete, or for those in Cluster 4, which consistently follow a two-block structure.
This observation is also supported by Figure~\ref{fig:tfr}, where we observe a 9.2\% increase in the TFR metric for the COLLAB dataset.


\vspace{-2mm}
\section{Conclusions}
\label{sec:concl}
\vspace{-2mm}
In this work, we introduced MGCL, a model-driven framework for contrastive learning on graphs. 
MGCL assumes that observed graphs are generated by an underlying generative process, modeled as a graphon. 
This perspective addresses a central challenge in GCL: the dependence on manually crafted augmentations. 
By estimating the graphon from data, MGCL constructs graphon-informed augmentations -- structure-aware stochastic transformations that reflect the generative process. 
This principled approach enables contrastive learning to align with the latent structure of the data, moving beyond individual graph instances.

A key limitation of our approach lies in the simplicity of the graphon model, which, while effective for augmentation, may not capture the full complexity of real-world graph generation, as it models only the graph structure and does not account for node or edge features.
To address this, future work can explore more expressive generative models, such as diffusion-based models, to further enhance the model-based contrastive learning framework.
Additionally, a natural extension is to utilize
InfoNCE-style objectives in node-level tasks, replacing the current reliance on DGI-based methods.

\newpage

\section*{Acknowledgement}
This research was sponsored by the Army Research Office under Grant Number W911NF-17-S-0002 and by the National Science Foundation under awards CCF-2340481 and EF-2126387. 
NZ was partially supported by a Ken Kennedy Institute 2024–25 Ken Kennedy-HPE Cray Graduate Fellowship.
The views and conclusions contained in this document are those of the authors and should not be interpreted as representing the official policies, either expressed or implied, of the Army Research Office, the U.S. Army, or the U.S. Government. 
The U.S. Government is authorized to reproduce and distribute reprints for Government purposes, notwithstanding any copyright notation herein.

\bibliographystyle{apalike}
\bibliography{MGCL}

\newpage
\appendix


\section{Algorithm}
\label{app:alg}

We present the pseudocode of the node-level MGCL in Alg.~\ref{alg:node-level}, while the graph-level case is in Alg.~\ref{alg:graph-level}.
Both algorithms rely on Alg.~\ref{alg:gia}, which implements the graphon-informed augmentation \(\ccalT_{\text{GIA}}\) introduced in Section~\ref{subsec:methodNode}.

\begin{algorithm}[!htb]
	\caption{Generate augmentations $\ccalT_{\mathrm{GIA}}(.)$}\label{alg:gia}
	\begin{algorithmic}[1]
 	\Require $\bbA, r, f_{\phi}, g_{\phi'}$
    \State \(\bbeta = g_{\phi'}(\bbA)\)
    \State Select subset of edges $E_{sel}$ with size $|E_{sel}| = (r/100)\times (n(n-1)/2)$
    \For{each pair $(i,j) \in E_{sel}$}
    \State $\tbA(i,j) \sim \mathrm{Bernoulli}\left(f_{\phi}(\bbeta(i), \bbeta(j))\right)$
    \State $\tbA(j,i) = \tbA(i,j)$
    \EndFor
    \State $\tbA(l,m) = \bbA(l,m)$ for $(l,m) \notin E_{sel}$ \\
    \Return $\tbA$
	\end{algorithmic}
\end{algorithm}
\begin{algorithm}[!htb]
	\caption{Node-level MGCL}\label{alg:node-level}
	\begin{algorithmic}[1]
 	\Require $\bbA, r, \bbX$\\
    \State $\verb|Step 1: Graphon estimation using SIGL|$
    \State Sample $\bbY \sim \ccalN(0, \bbI)$
    \State $f_{\phi}, g_{\phi'} = \mathrm{SIGL}(\bbA, \bbY)$\\
    \State $\verb|Step 2: Generate augmentations via|$ $\ccalT_{GIA}(.)$
    \State $\verb|Generate positive view|$ $\tbA_p$
    \State $(\tbA_p, \bbX) = \mathrm{Algorithm~\ref{alg:gia}}(\bbA, r, f_{\phi}, g_{\phi'})$
    \State $\verb|Generate negative view|$ $\tbA_n$
    \State $\tbX \leftarrow$ Shuffle $\bbX$
    \State $(\tbA_n, \tbX) = \mathrm{Algorithm~\ref{alg:gia}}(\bbA, r, f_{\phi}, g_{\phi'})$\\
    \State $\verb|Step 3: Train encoder|$
    \For{$l = 1,\cdots, L$}
    \State $\bbH = \ccalE_{\theta}(\bbA, \bbX)$
    \State $\tbH_n = \ccalE_{\theta}(\tbA_n, \tbX)$
    \State $\bbs_p = \ccalR(\ccalE_{\theta}(\tbA_p, \bbX))$
    \State $ \ccalL_{\text{DGI}} = \frac{1}{2N} \left( \sum_{i=1}^{N} \log \ccalD(\bbh_i, \bbs_p) + \log\left(1 - \ccalD(\tbh_i^n, \bbs_p)\right) \right)$
    \State $\theta = \mathrm{OptimizerStep(\ccalL_{\text{DGI}})}$
    \EndFor \\
	\Return $\ccalE_{\theta^*}(.)$
	\end{algorithmic}
\end{algorithm}
\begin{algorithm}[!htb]
	\caption{Graph-level MGCL}\label{alg:graph-level}
	\begin{algorithmic}[1]
 	\Require $\ccalD = \{\bbA_t\}_{t=1}^L, r, \{\bbX_t\}_{t=1}^L$\\
    \State $\verb|Step 1: Graph clustering |$
    \State $\bbz_t^{\mathrm{init}} = g_{\zeta}(\bbA_t, \bbX_t)\quad \forall t=1, \cdots, L$
    \State Run $\verb|K-Means|$ $(\bbz_t^{\mathrm{init}})$ to obtain $K$ clusters\\
    \State $\verb|Step 2: Graphon estimation using SIGL|$
    \For{$k \in K$}
    \State Select the $J$ graphs $\{\bbA_{idx(j)}^{(k)}\}_{j=1}^J$ from cluster $k$ closest to its center
    \State $\{\bbY_j^{(k)}\}_{j=1}^{J} \sim \ccalN(0, \bbI)$
    \State $f_{\phi}^{(k)}, g_{\phi'}^{(k)} = \mathrm{SIGL}(\{\bbA_{idx(j)}^{(k)}, \bbY_j^{(k)}\}_{j=1}^J)$
    \EndFor\\
    \State $\verb|Step 3: Generate augmentations|$
    \For{$t \in L$}
    \State Get cluster $c(t)$ of $\bbA_t$
    \State $(\tbA_t^1, \bbX_t) = \mathrm{Algorithm~\ref{alg:gia}}(\bbA_t, \bbX_t, f_{\phi}^{c(t)}, g_{\phi'}^{c(t)})$
    \State $(\tbA_t^2, \bbX_t) = \mathrm{Algorithm~\ref{alg:gia}}(\bbA_t, \bbX_t, f_{\phi}^{c(t)}, g_{\phi'}^{c(t)})$
    \EndFor\\
    \State $\verb|Step 4: Train encoder|$
    \For{$l \in L_{steps}$}
    \For{$t \in L$}
    \State $\bbz_t^1 = \ccalE_{\theta}(\tbA_t^1, \bbX_t)$ 
    \State $\bbz_t^2 = \ccalE_{\theta}(\tbA_t^2, \bbX_t)$
    \State $\ell_t = -\log \frac{\exp (\text{sim}(\bbz_t, \bbz_t^1) / \tau)}{\sum_{t'=1,\, c(t') \neq c(t)}^L \exp (\text{sim}(\bbz_t, \bbz_{t'}^2) / \tau)}$
    \EndFor
    \State $\ccalL_{\text{all}} = \frac{1}{L}\sum_{t=1}^{L}\ell_t$
    \State $\theta = \mathrm{OptimizerStep(\ccalL_{all})}$
    \EndFor\\
    \Return $\ccalE_{\theta^*}(.)$
	\end{algorithmic}
\end{algorithm}

\section{Graphon estimation}
\label{app:graphon}

The goal is to estimate an unknown graphon \( \omega: [0,1]^2 \to [0,1] \), given a set of graphs \( \ccalD = \{\bbA_t\}_{t=1}^{M} \) sampled from it. 
Since using the Gromov-Wasserstein (GW)~\cite{gw} distance is computationally infeasible for large graphs, the SIGL framework~\cite{sigl} proposes a scalable three-step procedure:

\paragraph{Step 1: Sorting nodes via latent variable estimation}

To align all graphs to a common node ordering (which is crucial for consistent estimation), SIGL estimates latent variables \( \hbeta_t = \{\hat{\eta}_i\}_{i=1}^{n_t} \) for each graph \( G_t \) using a Graph Neural Network (GNN):
\[
\hbeta_t = g_{\phi_1}(\bbA_t, \bbY_t), \quad \text{where } \bbY_t \sim \mathcal{N}(0,1)
\]

An auxiliary graphon \( h_{\phi_2} \) modeled by an Implicit Neural Representation (INR) maps pairs of latent variables to edge probabilities:
\[
h_{\phi_2}(\hbeta_t(i), \hbeta_t(j)) \approx \bbA_t(i,j)
\]

The latent variables and auxiliary graphon are jointly trained by minimizing the mean squared error to get \(\phi = \{\phi_1 \cup \phi_2\}\):
\[
\ccalL(\phi) = \sum_{t=1}^{M} \frac{1}{n_t^2} \sum_{i,j=1}^{n_t} \left[ \bbA_t(i,j) - h_{\phi_2}(\hbeta_t(i), \hbeta_t(j)) \right]^2
\]

A sorting permutation \( \hat{\pi} \) is defined based on the learned latent variables:
\[
\hbeta_t(\hat{\pi}(1)) \geq \hbeta_t(\hat{\pi}(2)) \geq \cdots \geq \hbeta_t(\hat{\pi}(n_t))
\]
In a nutshell, this permutation sorts the latent variables from $0$ to $1$. 
The graphs are reordered accordingly to produce sorted adjacency matrices \( \hat{\bbA}_t \).

\paragraph{Step 2: Histogram Approximation}

For each sorted graph \( \hat{\bbA}_t \), a histogram \( \hbH_t \in \mathbb{R}^{k \times k} \) is computed using average pooling with window size \( h \):
\[
\hbH_t(i,j) = \frac{1}{h^2} \sum_{s_1=1}^{h} \sum_{s_2=1}^{h} \hbA_t\left((i-1)h + s_1, (j-1)h + s_2\right)
\]

This results in a new dataset \( \ccalI = \{\hbH_t\}_{t=1}^{M} \), providing discrete, noisy views of the unknown graphon \( \omega \).

\paragraph{Step 3: Training the Graphon INR}

The final step constructs a supervised dataset \( \ccalC \) from all histograms, where each point corresponds to a coordinate-value triple:
\[
\ccalC = \left\{ \left( \frac{i}{k_t}, \frac{j}{k_t}, \hbH_t(i,j) \right) : i,j \in \{1,\dots,k_t\}, \, t \in \{1,\dots,M\} \right\}
\]

A second INR structure \( f_\theta: [0,1]^2 \to [0,1] \) is then trained to regress the graphon values by minimizing the MSE:
\[
\mathcal{L}(\theta) = \sum_{(x,y,z) \in \ccalC} \left( f_\theta(x,y) - z \right)^2
\]

This scalable approach enables two things: 1) the estimation of a continuous graphon \( \omega \) using large-scale graph data without relying on costly combinatorial metrics like the GW distance, and 2) the estimation of the latent variables given an input graphs, i.e., an inverse mapping $\ccalW^{-1}: \bbA \rightarrow \bbeta$.

\section{Experimental details}
\label{app:implementation}

\subsection{Hyper-parameter}
\label{subsec:hyper}
\paragraph{Graphon estimation} 
The hyperparameters used to estimate the graphon with SIGL across its three steps, as described in the previous section, are as follows for both node and graph level tasks. 
We use the \verb|Adam| optimizer~\cite{kingma2014adam} with a learning rate of $lr = 0.01$ for both Step 1 and Step 3, running for 40 and 20 epochs, respectively. 
In Step 1, the batch size is set to 1 graph, while in Step 3, each batch includes 1024 data points from \(\ccalC\). 
In Step 1, the GNN, $g_{\phi_1}$ comprises two consecutive graph convolutional layers, each followed by a ReLU activation function. 
All convolutional layers use 8 hidden channels. 
The INR structures in Step 1 ($h_{\phi_2}$) and Step 3 ($f_{\theta}$) each have 3 layers with 20 hidden units per layer and use a default frequency of 10 for the $\sin(.)$ activation function.

\paragraph{Node-level MGCL} 
After estimating the graphon for node-level tasks, we set \(r = 20\%\) and resample 20\% of the entries in the adjacency matrix of the graph. 
For training the encoder, we follow the setup of GraphCL~\cite{graphcl}, which is an adaptation of DGI~\cite{dgi}. 
We use the \verb|Adam| optimizer with a learning rate of \(lr = 0.001\). 
Training is stopped if the loss does not decrease for 20 consecutive epochs. 
The dimension of the learned representations is 512. 
The encoder \(\ccalE_\theta\) is structured as a single-layer GCN followed by a ReLU activation. 
The discriminator function is defined as a bilinear function \(\ccalD(\bbh, \bbs) = \sigma(\bbs^\top W \bbh)\), where \(W\) is the learnable weights.
For the readout function \(\ccalR(.)\), we use average pooling over node embeddings.

\paragraph{Graph-level MGCL}

To obtain \(\bbz_t^{\text{init}}\) and cluster the graphs, we use a randomly initialized GNN \(g_{\zeta}(\cdot)\), structured as a 3-layer GCN with 32 hidden units in each layer. 
The dimension of the initial representations \(\bbz_t^{\text{init}}\) is set to 32. 
To focus solely on the graph structure and avoid reliance on node features in estimating the graphons, we assign an all-ones vector to each node in this step.
After clustering the graphs into \(K = \log(L)\) groups, we select the \(J=10\) graphs closest to the center of each cluster to estimate the graphon for that cluster. 
This selection aims to reduce the effect of noisy samples and focus on the central and shared structures within each group, avoiding graphs that lie on the boundary between clusters.

To train the encoder \(\ccalE_\theta\), we follow the configuration from GraphCL~\cite{graphcl}. 
We use the \verb|Adam| optimizer with a learning rate of \(lr = 0.001\), training the encoder for 20 epochs. 
The encoder is implemented as a 3-layer GIN~\cite{gin} network, with each layer consisting of 32 hidden units followed by a ReLU activation. 
A final linear projection head maps the output to a 32-dimensional graph-level representation.

\subsection{Synthetic node classification details}
Here, we describe the experiment "Effect of Graphon-Informed Augmentation (GIA)" presented in Section~\ref{result_real}. 
This setup models a scenario in which the graph is generated from a latent graphon model, and node classes depend on their underlying latent variables.
We consider the ground-truth graphon to be \(\ccalW(x, y) = xy\). 
Using this graphon, we construct a graphon neural network (WNN)~\cite{graphonwnn} with two layers, containing 4 and 1 features respectively, where the filter coefficients in each layer are randomly selected. 
Each layer in the WNN can be interpreted as a continuous analogue of a graph convolution layer.
This WNN serves as a non-linear function that maps each latent variable \(v\) in the graphon domain to an output value, taking into account the underlying graphon structure:\(v \stackrel{\ccalW}{\rightarrow} o \in [0,1]\).
To obtain binary labels, we threshold the WNN output using the average value across all latent variables. 

\begin{figure}[h]
    \centering
    \includegraphics[width=0.5\textwidth]{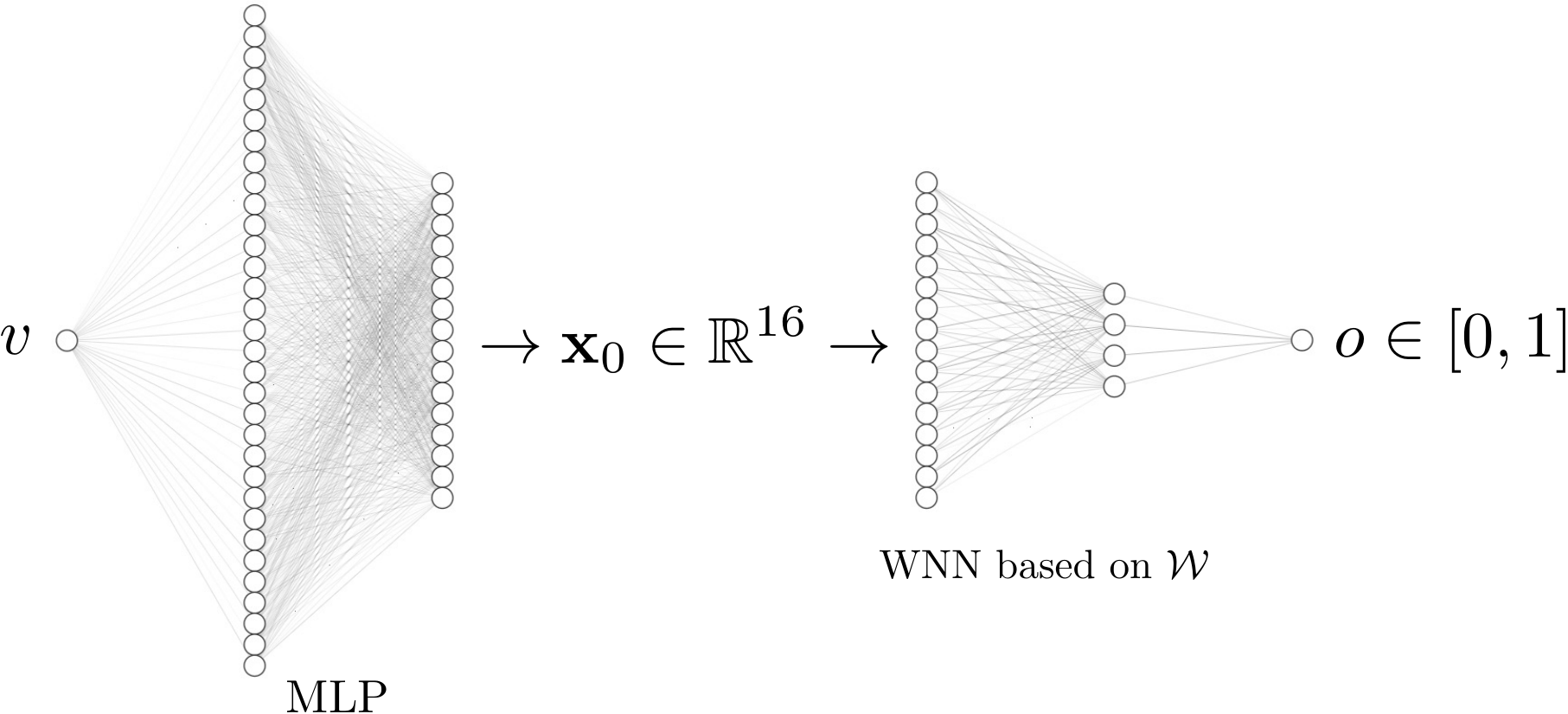}
    \caption{Simulating a graph, its initial features, and node labels using a graphon neural network.}
    \label{fig:wnn}
\end{figure}

For each trial, we randomly sample a graph with a size between 300 and 700 nodes, with known latent variables for each node. 
The label of each node is determined based on the WNN output associated with its latent variable.
To generate initial features for each node, we use a randomly initialized multi-layer perceptron (MLP) with two layers of 32 and 16 hidden units, respectively, which maps each latent variable \(v\) to a 16-dimensional feature vector in a non-linear fashion.
Following these steps, we obtain a graph with node features and labels derived from a coherent generative model. 
All subsequent steps in the pipeline mirror those used for real-world benchmark datasets.
An overview of the process for generating the graph, initial features, and output labels is illustrated in Figure~\ref{fig:wnn}.

\subsection{Dataset Details}

For node-level experiments, we use six benchmark datasets:

\begin{itemize}
    \item \textbf{Citation networks}~\cite{sen2008collective}: Cora, Citeseer, and PubMed. In these datasets, each node represents a scientific publication, and edges denote citation relationships between them. Node features are derived from document attributes such as abstracts, keywords, or full text. The labels indicate the research topic or category of each paper.
    
    \item \textbf{(Amazon-)Photo}~\cite{shchur2018pitfalls}: This dataset captures the co-purchase relationships between products on Amazon. Nodes represent products, and edges connect items frequently bought together. Products are divided into eight categories, and node features are encoded using a bag-of-words representation based on product reviews.
    
    \item \textbf{(Coauthor-)CS and (Coauthor-)Physics}~\cite{shchur2018pitfalls}: These datasets are derived from the Microsoft Academic Graph and model co-authorship relationships among researchers. Nodes correspond to authors, and edges indicate collaborative publications. Authors are labeled according to their research fields (15 for CS and 5 for Physics), with features represented as a bag-of-words vector of publication keywords.
\end{itemize}

The detailed statistics for these datasets are presented in Table~\ref{tab:node_data}.

\begin{table*}[h]
\centering
\caption{Node-level dataset benchmarks statistics.}
\label{tab:node_data}
\resizebox{0.7\textwidth}{!}{
\begin{tabular}{l|cccccc}
\toprule
{Statistic} & {Cora} & {CiteSeer} & {PubMed} & {Photo} & {Cs} & {Physics} \\
\midrule
\# Nodes & 2,708 & 3,327 & 19,717 & 7,650 & 18,333 & 34,493 \\
\# Edges & 5,429 & 4,732 & 44,338 & 119,081 & 81,894 & 991,848 \\
\# Features & 1,433 & 3,703 & 500 & 745 & 6,805 & 8,451 \\
\# Classes & 7 & 6 & 3 & 8 & 15 & 5 \\
\bottomrule
\end{tabular}
}
\end{table*}

For graph-level tasks, we use datasets from the TUdataset collection, which includes biochemical molecule graphs and social network graphs. Detailed statistics of these datasets are provided in Table~\ref{tab:graph_data}.

\begin{table*}[h]
\centering
\caption{Graph-level dataset benchmarks statistics.}
\label{tab:graph_data}
\resizebox{\textwidth}{!}{
\begin{tabular}{l|cccc|cccc}
\toprule
 & \multicolumn{4}{c|}{{Biochemical Molecules}} & \multicolumn{4}{c}{{Social Networks}} \\
{Statistic} & {NCI1} & {PROTEINS} & {DD} & {MUTAG} & {COLLAB} & {RDT-B} & {RDT-M5K} & {IMDB-B} \\
\midrule
\#Graphs & 4,110 & 1,113 & 1,178 & 188 & 5,000 & 2,000 & 4,999 & 1,000 \\
Avg. \#Nodes & 29.87 & 39.06 & 284.32 & 17.93 & 74.5 & 429.6 & 508.8 & 19.8 \\
Avg. \#Edges & 32.30 & 72.82 & 715.66 & 19.79 & 2457.78 & 497.75 & 594.87 & 96.53 \\
\#Classes & 2 & 2 & 2 & 2 & 3 & 2 & 5 & 2 \\
\bottomrule
\end{tabular}
}
\end{table*}

Note that several graph-level datasets lack explicit node attributes. In such cases, one-hot encoding of node degrees is commonly used to construct node features.

\subsection{Baseline models.}

An introduction to the baselines for both node-level and graph-level tasks is outlined below. 

GCN~\cite{gcn}: A representative Graph Neural Network (GNN) that combines spectral and spatial strategies to perform graph convolution. 
It enables each node to aggregate information from its neighbors by incorporating both the graph topology and node attributes. 

DGI~\cite{dgi}: A GCL model based on the Infomax principle, which augments the graph by row-wise shuffling of the attribute matrix and maximizes the mutual information between global and local representations.

MVGRL~\cite{mvgrl}: Another DGI variant that performs contrastive learning between different structural views of graphs, such as first-order adjacency and graph diffusion.

GRACE~\cite{grace}: A GCL model that generates node embeddings by corrupting the graph structure (via random edge removal) and node attributes (via random masking) to produce diverse views and maximize their agreement.

GCA~\cite{gca}: A variant of GRACE that introduces adaptive augmentation strategies based on node and edge centrality to improve model flexibility.

GraphCL~\cite{graphcl}: A GCL framework that learns graph representations by applying various augmentations to local subgraphs of nodes.

JOAO~\cite{joao}: A variant of GraphCL that employs min-max optimization to automatically select the most effective augmentations during contrastive learning.

BGRL~\cite{bgrl}: A GCL model that applies node feature masking and edge masking for graph augmentation and uses bootstrapping to update the online encoder's parameters.

GBT~\cite{gbt}: A feature-level GCL model that uses the Barlow Twins loss to reduce redundancy between two augmented views created through random edge removal.

InfoGraph~\cite{infograph}: A variant of DGI that maximizes mutual information between graph-level representations and substructures at different scales, including nodes, edges, and triangles.

\subsection{Compute resources}

All experiments were conducted on a server running Ubuntu 20.04.6 LTS, equipped with an AMD EPYC 7742 64-Core Processor and an NVIDIA A100-SXM4-80GB GPU with 80GB of memory.
For model development, we utilized PyTorch version 1.13.1, along with PyTorch Geometric version 2.3.1, which also served as the source for all datasets used in our study.

\section{Additional experiments}
\label{app:experiments}

\subsection{Effect of resampling ratio in node-level task}
\begin{figure}[h]
    \centering
    \includegraphics[width=0.70\textwidth]{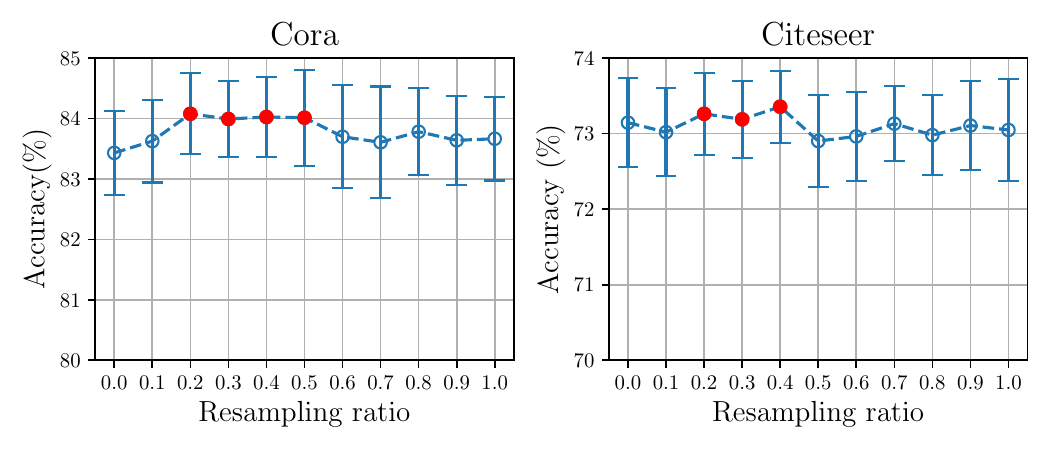}
    \caption{{Effect of graphon-based resampling ratio \(r\) on classification accuracy.}}
    \label{fig:rp_effect}
\end{figure}

In this experiment, we study the impact of the resampling ratio $r$ in the \(\ccalT_{\text{GIA}}\) on node classification performance.
We vary \(r\) from 0.0 to 1.0 in increments of 0.1 on two datasets, Cora and CiteSeer, and report the average classification accuracy over 10 trials using fixed data partitions across all settings.
Note that when \(r = 0.0\), the positive view is identical to the original graph, and the negative view differs only in node features (which are permuted).
At the other extreme, when \(r = 1.0\), the entire graph structure is resampled from the estimated graphon. 
As depicted in Figure~\ref{fig:rp_effect}, the results reveal an optimal performance range for the resampling ratio, with MGCL achieving peak accuracy on both datasets when \(r\) lies approximately between 0.2 and 0.4.
This suggests that partial resampling introduces sufficient perturbation to benefit contrastive learning without deviating too far from the original graph structure.


\subsection{Number of clusters in graph-level tasks}

As mentioned in Section~\ref{subsec:methodGraph}, for graph-level tasks we cluster the dataset into \(K = \log(L)\) groups, where \(L\) is the number of graphs. 
In this section, we vary the number of clusters to evaluate its impact on overall performance.

We vary the number of clusters from 1 to 10. 
For each value, we repeat the graph classification experiment across the same 10 trials using identical data partitions and report the average accuracy.

\begin{figure}[h]
    \centering
    \includegraphics[width=0.90\textwidth]{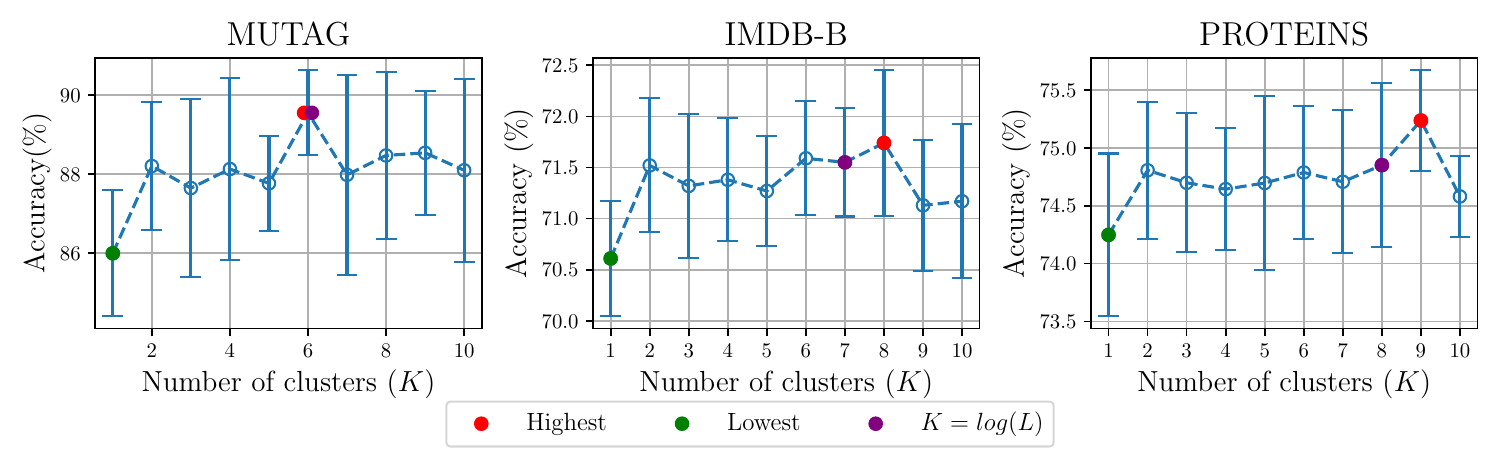}
    \caption{{Effect of the number of clusters on graph classification performance.}}
    \label{fig:cluster}
\end{figure}

The results are presented in Figure~\ref{fig:cluster} for the MUTAG, IMDB-BINARY, and PROTEINS datasets. 
A key observation is that using a single cluster—i.e., estimating one graphon for the entire dataset—leads to a drop in performance across all three datasets. 
In the IMDB-BINARY dataset, using 8 clusters yields better performance than the default setting of 7 clusters. 
However, using between 6 and 8 clusters consistently results in an average accuracy above 71.5\%, suggesting that this range reasonably approximates the number of underlying models.
A similar trend is observed in the PROTEINS dataset, where 9 clusters yield better performance than 8. 
For the MUTAG dataset, the highest average accuracy is achieved with the default setting of 6 clusters. 
These findings highlight that estimating multiple models improves performance. Still, the number of clusters remains a hyperparameter that should be tuned based on the dataset’s characteristics, such as its variability and heterogeneity.

\subsection{Effect of having two augmentations in graph-level tasks}
As mentioned in~\ref{subsec:methodGraph}, for each graph, we generate two augmentations based on its corresponding model. 
We then encourage the graph representation to align closely with its first augmentation (positive view) while being dissimilar from the second augmentations of all other graphs in different clusters (negative views). 

To evaluate the effect of using two augmentations, we conduct an experiment in which only a single augmentation is generated for each graph. 
This single augmentation serves as both the positive view for the original graph and as the negative counterpart for all other graphs in different clusters. 
We refer to this variant as {MGCL-1}. 
We repeat the graph classification experiment over the same 10 trials used in the main setting and report the average accuracy.

\begin{table*}[h]
\centering
\caption{Performance comparison between using two augmentations (MGCL) vs. a single augmentation (MGCL-1).}
\label{tab:oneAug}
\resizebox{1\textwidth}{!}{
\begin{tabular}{l|cccc|cccc}
\toprule
{Method} & {NCI1} & {PROTEINS} & {DD} & {MUTAG} & {COLLAB} & {RDT-B} & {RDT-M5K} & {IMDB-B} \\
\midrule
MGCL & 78.66{\scriptsize$\pm$0.34} & 74.85{\scriptsize$\pm$0.71} & 78.88{\scriptsize$\pm$0.38} & 89.55{\scriptsize$\pm$1.08} & 71.44{\scriptsize$\pm$0.71} & 90.25{\scriptsize$\pm$0.39} & 55.65{\scriptsize$\pm$0.32} & 71.55{\scriptsize$\pm$0.53} \\
MGCL-1 & 78.50{\scriptsize$\pm$0.63} & 74.36{\scriptsize$\pm$0.60} & 78.66{\scriptsize$\pm$0.99} & 87.66{\scriptsize$\pm$1.73} & 70.99{\scriptsize$\pm$0.64} & 90.05{\scriptsize$\pm$0.43} & 55.37{\scriptsize$\pm$0.34} & 71.24{\scriptsize$\pm$0.62} \\
\bottomrule
\end{tabular}
}
\end{table*}

In Table~\ref{tab:oneAug}, we compare the performance of MGCL-1 with the standard MGCL.
We observe a decrease in average accuracy across all datasets when using a single augmentation.
This indicates that generating two augmentations per graph enables the model to learn more discriminative representations by enhancing the model-aware contrastive signal.



\subsection{Graphon Estimation in MGCL}

We evaluate the effectiveness of our graphon estimation procedure described in Section~\ref{subsec:methodGraph}. 
To this end, we construct a synthetic dataset by sampling 250 graphs from each of four distinct ground-truth graphons, resulting in a total of 1000 graphs. 
We then apply the first two steps of the graph-level MGCL methodology to cluster these graphs and estimate a separate graphon for each resulting cluster.

Our goal is twofold: (i) to assess whether graphs generated from the same underlying graphon are correctly grouped into the same cluster, and (ii) to evaluate how well each estimated graphon matches its corresponding ground-truth graphon.

\begin{figure*}[h]
	\centering
	\includegraphics[width=0.85\textwidth]{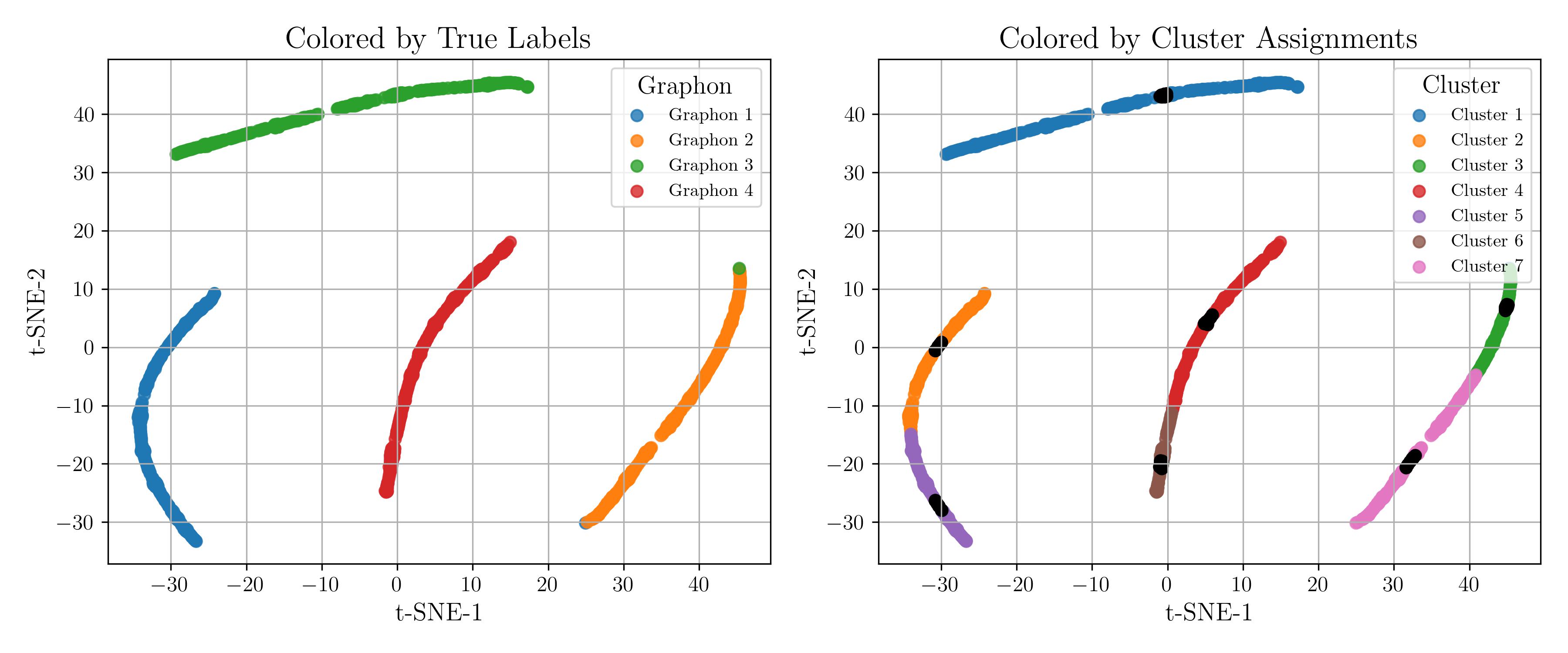}
	\caption{Initial graph embeddings colored by true graphon (left) and MGCL cluster assignments (right).}
	\label{fig:cluster_sim}
\end{figure*}  

In Figure~\ref{fig:cluster_sim}, we visualize the initial GNN embeddings of the graphs (\(\bbz^{\text{init}}\)). 
The left plot colors the embeddings based on their true generating graphon, while the right plot shows the cluster assignments produced by MGCL. 
We observe that graphs originating from different graphons are well-separated in the embedding space. 
Three of the true graphon groups are further divided into two subclusters. 
This behavior is expected, as MGCL clusters the graphs into \(K = \log(L)\) groups—resulting in 7 clusters for \(L = 1000\)—which produces finer partitions than the actual number of ground-truth graphons.

\begin{figure*}[h]
	\centering
	\includegraphics[width=0.95\textwidth]{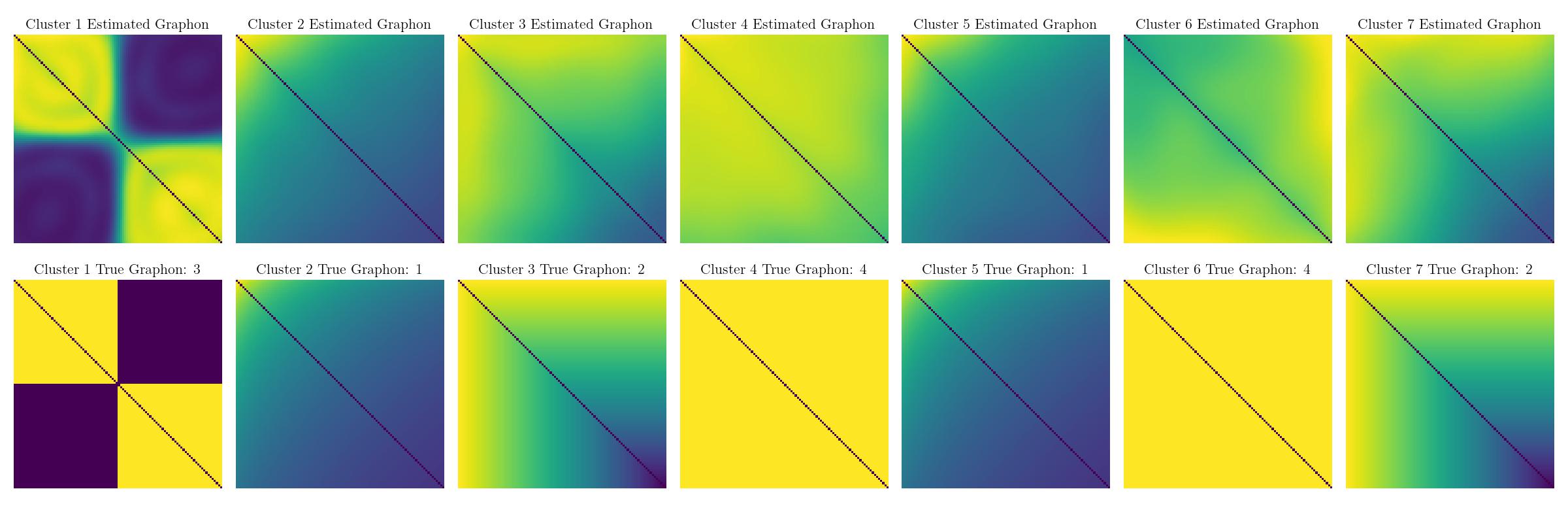}
	\caption{Estimated graphons compared to the ground-truth graphons.}
	\label{fig:graphon_sim}
\end{figure*}

Moreover, in Figure~\ref{fig:graphon_sim}, we compare each estimated graphon with its corresponding ground-truth graphon. 
The visual similarity between them indicates that MGCL is able to accurately estimate the underlying generative structures.

\end{document}